\documentclass[runningheads]{llncs}
\usepackage[T1]{fontenc}
\usepackage{hyphenat}
\usepackage{booktabs}
\usepackage{graphicx}
\usepackage{multirow}
\usepackage{hyperref}
\usepackage{color}

\usepackage{microtype}
\begin{document}
%
\title{
From Presence Labels to Bounding Boxes: Can Multimodal Large Language Models Scale Surgical Instrument Localization?}

\titlerunning{MLLMs for Scalable Surgical Instrument Annotation}
%

\author{Mohamed Hamdy\inst{1} \and
Muraam Abdel-Ghani\inst{2} \and
Fatmaelzahraa Ahmed\inst{2} \and
Sifna Nasar\inst{1} \and
Mariam Ahmed\inst{3} \and
Khalid Al-Jalham\inst{2} \and
Abdulaziz Al-Ali\inst{1} \and
Shidin Balakrishnan \thanks{Corresponding author: SBalakrishnan1@hamad.qa}\inst{2}}
\authorrunning{M. Hamdy et al.}
%
\institute{Department of Computer Science and Engineering, Qatar University, Doha, Qatar\and
Hamad Medical Corporation, Doha, Qatar\\
\and
College of Medicine, Qatar University, Doha, Qatar
}
\maketitle              
%

%
%
%

\begin{abstract}
Spatial localization of surgical instruments through bounding-box annotations is a foundational component of surgical scene understanding, supporting detection, tracking, interaction analysis, and grounded vision--language applications. Yet existing annotations often cover selected frames, video subsets, or dataset variants rather than complete procedures. Manual extension is time-consuming and expert-dependent. Additionally, weakly supervised and detector-based pseudo-labels require task-specific training and may be imprecise. In this study, we design a scalable MLLM-based annotation framework that converts surgical frames and existing instrument-presence labels into accurate bounding boxes, and systematically analyze its reliability, cost, and design sensitivity. We instantiate the framework using six MLLMs and a multi-call configuration on laparoscopic cholecystectomy in Cholec80. Using a carefully designed stratification scheme, two annotators independently label 1,000 frames, and an expert surgeon reviews one annotation set to establish the reference. Results show that instrument-set F1 ranges from 88.2 to 92.7, whereas mIoU ranges from 0.180 to 0.735, showing that comparable recognition can conceal large localization differences. A structured multi-stage annotation pipeline achieves 0.784 mIoU, compared with 0.735 for the best single-call baseline and 0.814 human--human agreement.
Using the best accuracy pipeline, we create and publicly release a spatial extension of Cholec80 comprising 285,977 bounding boxes across all 184,498 annotated frames from the 80 videos. The annotations and human references are made available on GitHub\footnote{\url{https://github.com/M-Hamdy-M/SurgBox}}.

\end{abstract}

\section{Introduction}
Minimally invasive surgery (MIS) has become widely adopted over the past decades due to its association with faster recovery and reduced intraoperative blood loss \cite{MIS}. The endoscopic cameras utilized in MIS generate continuous video documentation of anatomy, tools, activities, and workflow, supplying extensive data for computer vision systems \cite{systematic_review,hamdy2026decoupling,midlshort}. Models have advanced from coarse phase and instrument detection \cite{twinanda2017endonet} to fine-grained tasks like instrument detection, segmentation \cite{brdg,CLIP-RL,fasl}, and action triplet recognition \cite{nwoye2020tripnet,nwoye2022rendezvous}. Recently, large vision-language models have enabled surgical visual question answering and multimodal scene understanding \cite{wang2024surgicallvlm,zeng2025surgvlm}. However, these detailed tasks often require spatially grounded annotations, which are much harder and more expensive to obtain than simple frame-level labels.

The availability of spatial supervision has not kept pace with the expansion of surgical video understanding. In laparoscopic cholecystectomy, Cholec80 provides phase and instrument-presence annotations yet  instance-level localization is absent ~\cite{twinanda2017endonet}, and its widely used VQA extension, Cholec80-VQA, similarly contains question-answer supervision without visual grounding~\cite{surgicalvqa}. This limitation persists in recent multimodal resources. although Surg-396K introduced grounded surgical conversations through EndoChat, its Cholec80-derived component inherits the non-grounded Cholec80-VQA annotation. Therefore it is used only for non-spatial question answering~\cite{endochat}. Dedicated localization datasets provide stronger spatial supervision but remain limited in coverage or annotation granularity. CholecT50 provides instrument-tip bounding boxes for five videos and six instrument classes~\cite{cholectriplet2022}, leaving its spatial coverage more limited than the full Cholec80 dataset. Endoscapes provides bounding-box annotations for 1,933 sparsely sampled frames drawn from the critical view of safety (CVS) regions of 201  MIS cholecystectomy videos, with instruments represented as a single generic tool class \cite{murali2023endoscapes}. Thus, existing resources support either large-scale semantic supervision or small-scale spatial supervision, leaving a gap in dense, instance-level localization.

We therefore investigate whether current MLLMs can generate accurate instrument bounding boxes under these conditions, whether semantic recognition is indicative of localization quality, and whether decomposing the annotation task into simpler stages improves reliability.

To address these questions, we construct a 1,000-frame human reference spanning all 80 Cholec80 procedures. Each frame is independently annotated by two annotators, and one annotation set is subsequently reviewed by an expert surgeon. Using this reference, we evaluate the ability of six contemporary MLLMs to transform Cholec80's frame-level instrument-presence labels into class-specific, instance-level bounding boxes. We compare a direct single-call protocol with a structured workflow that separates instrument-set identification, instance counting, localization, and bounding-box refinement.

Our main contributions are:

 \begin{itemize}
     \item We construct a validated 1,000-frame human reference spanning all 80 Cholec80 videos, with two independent annotation passes and expert-surgeon review, enabling measurement of human--human spatial agreement.
     \item We benchmark six contemporary MLLMs for converting weak frame-level instrument supervision into class-specific instance-level bounding boxes.
     \item We develop a structured multi-stage annotation workflow and release SurgBox, containing 285,977 bounding boxes over all 184,498 annotated Cholec80 frames.
     \item We demonstrate the \textbf{downstream utility and transferability of SurgBox} by benchmarking eight detectors trained exclusively on MLLM-generated boxes and evaluating cross-dataset generalization to independently annotated CholecSeg8k, where SurgBox supervision achieves 0.833 mAP and reaches 0.866 mAP with subsequent in-domain fine-tuning.
 \end{itemize}


\section{MLLM-Based Surgical Instrument Annotation}
\label{sec:method}
We consider the problem of upgrading existing frame-level instrument-presence annotations into instance-level spatial supervision. The data source is from data Cholec80, for which binary presence labels are provided for seven instrument classes at 1\,fps, yielding 184,498 annotated frames. These labels indicate whether an instrument class is present but provide neither instance counts nor spatial locations. For each annotated frame, our objective is to infer the set of visible instrument instances and assign each a class-specific bounding box. Bounding boxes are defined over the currently visible extent of the instrument, including the shaft and working tip, without a modal completion of occluded or out-of-frame regions.

\begin{figure*}[t]
\centering
\includegraphics[width=\textwidth]{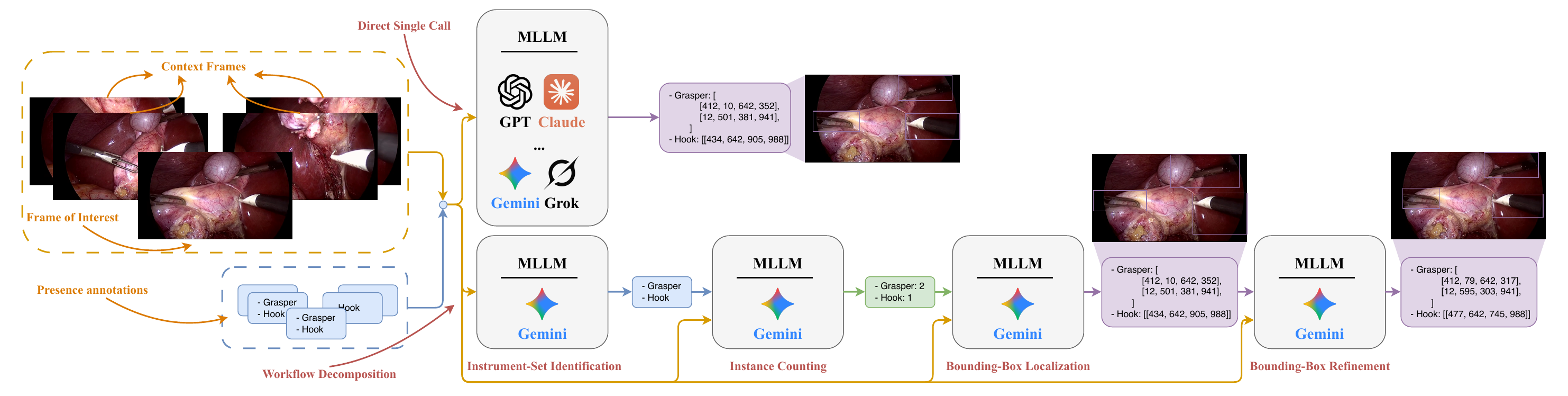}
\caption{Direct single-call and structured multi-stage workflows for MLLM-based surgical instrument bounding-box generation.}
\label{fig:annotation_workflows}
\end{figure*}

We formulate this conversion as an MLLM-based annotation workflow that combines the target frame with temporal visual context, the corresponding local window of Cholec80 presence labels, and textual descriptions of the instrument classes. 
Given this multimodal context $\mathcal{X}_t$, the workflow produces
an instance-level annotation
$\mathcal{B}_t=\{(c_i,b_i)\}_{i=1}^{N_t}$,
where $c_i$ denotes
the instrument class and $b_i$ its bounding box in the target frame. The original labels are treated as weak supervision, additionally, to prevent unconstrained corrections, the predicted instrument set is restricted to at most one modification: either a single addition or a single removal, with any added instrument required to appear within the nearby temporal labels. Under this formulation, we investigate two workflows: direct bounding-box generation in a single MLLM call, and a structured multi-stage workflow that separates instrument-set identification, instance counting, localization, and bounding-box refinement.
Figure~\ref{fig:annotation_workflows} summarizes these two annotation workflows.

\subsection{Direct Bounding-Box Generation}

First we consider a direct annotation workflow where  MLLM calls jointly identify, count, and localize the instruments visible in the target frame. The model receives the target frame, neighboring frames from the same video to provide short-term temporal information for resolving transient visibility changes and ambiguous appearances, the instrument-presence labels from a nearby temporal interval to provide weak procedural guidance on which tools are likely to be present, and textual descriptions of instrument classes to provide domain-specific visual cues. Given this multimodal input $\mathcal{X}_t$, the model directly predicts an instance-level annotation $\mathcal{B}_t=\{(c_i,b_i,s_i)\}_{i=1}^{N_t}$, where $c_i$ denotes the instrument class, $b_i=[y_{\min},x_{\min},y_{\max},x_{\max}]$ represents its bounding box on a normalized coordinate grid, and $s_i$ is the predicted confidence. Multiple instances of the same instrument class are allowed when several physical tools are visible. In our experiments, we instantiate this workflow using three frames from each video (the target frame and its immediate temporal neighbors), a five-point temporal label interval, and the seven Cholec80 instrument classes. This formulation evaluates whether an MLLM can perform the complete weak-to-spatial annotation process within a single prediction.

\subsection{Structured Multi-Stage Annotation}

The direct workflow requires a single MLLM call to simultaneously determine which instruments are visible and where each instance is located. These decisions involve different levels of visual reasoning and may interfere with each other, particularly when the model must distinguish visually similar instruments or localize multiple instances in cluttered surgical scenes. 

The first stage identifies the set of instruments visible in the target frame while validating the weak Cholec80 label. The second stage estimates the number of physical instances for each identified instrument class. This is necessary since the original binary labels cannot represent multiple occurrences of the same tool. Conditioned on the predicted instrument set and instance counts, the third stage performs localization by generating the corresponding bounding boxes. Finally, a refinement stage adjusts the predicted coordinates to improve box tightness while preserving the instrument identities and instance counts produced by the previous stages. 
All stages receive the same multimodal information as the direct workflow, while only the structured output of each preceding stage is passed forward. This design allows us to evaluate whether explicit task decomposition improves MLLM-based surgical annotation quality without introducing additional external supervision.

\subsection{Evaluation Framework}

To evaluate MLLM-generated annotations, we construct a human reference set of 1,000 frames from the 184,498 Cholec80 annotated frames. Since the natural distribution is highly imbalanced across instrument classes, we use a rarity-aware sampling strategy rather than uniform sampling. Specifically, frames are first divided into tool-present and tool-absent subsets, with 150 frames sampled from the latter to represent true negative cases. The remaining 850 frames are selected through an iterative greedy balancing procedure. At each iteration, a random candidate pool of 64 frames is sampled, and the selected frame is the one whose present instrument classes are most underrepresented relative to the target quota of approximately 143 occurrences per class. This process continues until the seven instruments achieve balanced representation within the evaluation subset.

Each selected frame is independently annotated by two annotators using a bounding-box protocol consistent with model predictions: boxes tightly enclose only the visible extent of each instrument, including the shaft and working tip, while excluding occluded or out-of-frame portions. One annotation pass is subsequently reviewed by an expert surgeon, and the reviewed annotations are used as the primary reference. The second independent annotation pass is retained to quantify human--human agreement and estimate the practical upper bound of spatial annotation consistency. This protocol provides both a high-quality evaluation reference and a direct measure of the intrinsic difficulty of surgical instrument localization.



\section{Experiments and Results}

\subsection{Experimental Setup}

\textbf{Dataset and evaluation protocol.}
We evaluate all approaches on Cholec80 with frame-level instrument-presence annotations for seven instrument classes: Grasper, Bipolar, Hook, Scissors, Clipper, Irrigator, and SpecimenBag. Model evaluation is performed on the 1,000-frame human reference set described in Section~\ref{sec:method}, while the full dataset is used only for large-scale generation after selecting the final annotation configuration.

\textbf{Compared models and annotation workflows.}
We benchmark six multimodal large language models from different model families, including Gemini, Claude, Grok, Llama, and Gemma variants. Each model is evaluated using the direct single-call annotation workflow under the same input formulation. We additionally evaluate the proposed structured multi-stage workflow using Gemini 3.1 Pro, selected on the basis of its strongest overall localization performance among the evaluated single-call models. This provides a strong reference point for isolating the effect of workflow decomposition.

\subsection{Results}
\textbf{Human Agreement and Reference Reliability. }
Before assessing model performance, we quantify agreement between the two independent annotators to establish the level of variability inherent to the annotation task. The two annotation sets achieve 94.18\% instrument-set F1 and 0.814 mIoU, increasing to 0.912 mIoU when computed only over matched boxes. Agreement is high for most instrument classes, while Grasper is substantially more difficult, with a human--human mIoU of 0.654 compared with 0.832--0.924 for the remaining classes. This human agreement provides a practical reference point for interpreting the localization accuracy of the evaluated MLLMs.


\textbf{Base-Model Capabilities. }
We first evaluate the native annotation capability of each MLLM using the direct single-call workflow, with the expert-reviewed primary annotations serving as the reference. Table~\ref{tab:base_models} summarizes semantic recognition, instance counting, localization, and inference cost across the six evaluated models. Gemini 3.1 Pro and Gemini 3.5 Flash consistently achieve the strongest performance across recognition, counting, and localization, with both reaching approximately 0.73 mIoU, while Gemma 4 shows moderate localization performance and Llama, Grok, and Claude exhibit substantially lower spatial accuracy. Inference cost varies by more than two orders of magnitude and shows no corresponding relationship with annotation quality; notably, the most expensive evaluated model is among the weaker localizers.  Gemini 3.1 Pro provides the strongest performance and is therefore selected for the subsequent structured workflow experiments.

\begin{figure*}[t]
\centering
\includegraphics[width=0.80\textwidth]{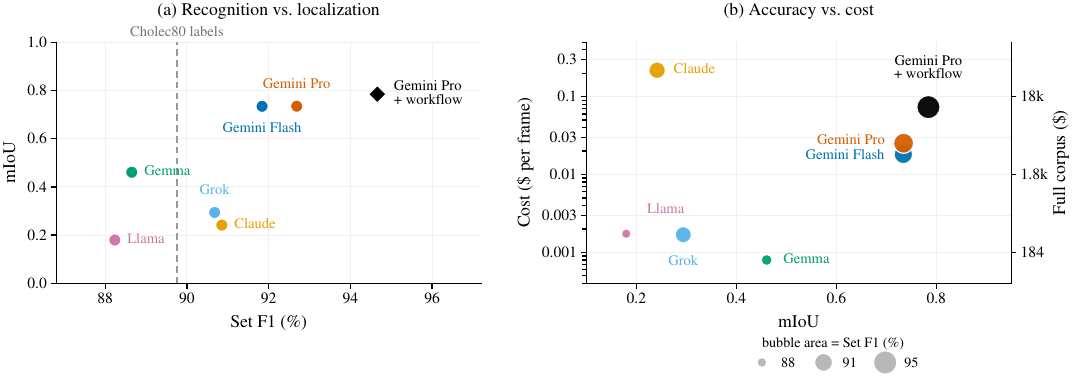}
\caption{Recognition, localization, and annotation-cost trade-offs. (a) Instrument-set F1 versus mIoU for the evaluated direct MLLMs and the selected structured workflow; the dashed line denotes the original Cholec80 label performance (89.75\% set F1). (b) Localization accuracy versus annotation cost per frame, with the right axis showing the corresponding linear cost projection over all 184,498 Cholec80 frames. Bubble area represents set F1.}
\label{fig:tradeoff}
\end{figure*}

Table~\ref{tab:base_models} shows that the original Cholec80 presence labels achieve 89.75\% set F1 against the human reference, providing all models with a strong semantic prior but no spatial information. Additionally, the table shows that permitting limited corrections enables the MLLMs to improve upon this prior rather than merely reproduce it: four of the six evaluated models exceed the original-label baseline, with Gemini 3.1 Pro reaching 92.69\% set F1, a gain of 2.94 percentage points compared with the original Cholec80 presence labels. Despite this relatively narrow range in semantic performance, localization varies substantially across models. This distinction is particularly evident for Claude Opus 4.8 and Grok 4.20, which exceed 90\% set F1 but achieve only 0.241 and 0.294 mIoU, respectively, compared with approximately 0.73 mIoU for the Gemini models. The disparity persists even among successfully matched instances, as reflected by the large variation in matched mIoU. Figure~\ref{fig:tradeoff}(a) visualizes this separation, demonstrating that accurate instrument-set prediction under weak-label conditioning does not necessarily translate into accurate spatial grounding.

\textbf{Structured Pipeline. }
Having identified Gemini 3.1 Pro as the strongest direct model for localization, we next evaluate the structured workflow described in Section~\ref{sec:method}. As shown in Table~\ref{tab:base_models}, decomposing instrument-set identification, instance counting, localization, and bounding-box refinement increases set F1 from 92.69\% to 94.66\% (+4.91 percentage points over the original Cholec80 labels), count exact match from 84.39\% to 88.24\%, and mIoU from 0.735 to 0.784, corresponding to 96.3\% of the observed human--human mIoU. Figure~\ref{fig:tradeoff}(a) further illustrates that the workflow improves both semantic recognition and spatial localization beyond direct generation.

\begin{table*}[t]
\centering
\caption{Performance on the 1,000-frame reference set. Parentheses in Set F1 report the percentage-point change relative to the original Cholec80 labels (89.75\%), while parentheses in mIoU report performance relative to human--human agreement (0.814). EM = exact match; Set = instrument-set identification; Cnt = instance counting; Loc = localization; Ref = bounding-box refinement.}
\label{tab:base_models}

\resizebox{\textwidth}{!}{
\begin{tabular}{ll|cccc|cc}
\toprule
\multicolumn{2}{c|}{} &
\multicolumn{4}{c|}{Quality} &
\multicolumn{2}{c}{Efficiency} \\
\cmidrule(lr){3-6}
\cmidrule(lr){7-8}
Model & Workflow &
\shortstack{Set F1\\(\%) $\uparrow$} &
\shortstack{Count EM\\(\%) $\uparrow$} &
\shortstack{mIoU\\$\uparrow$} &
\shortstack{Matched\\mIoU $\uparrow$} &
\shortstack{Calls\\per frame $\downarrow$} &
\shortstack{Cost\\(\$/frame) $\downarrow$} \\
\midrule

Cholec80 labels & -- &
89.75 (baseline) & -- & -- & -- & -- & -- \\

\midrule

Llama 4 Maverick & Direct &
88.24 ($-$1.51 pp) & 73.32 & 0.180 (22.1\%) & 0.215 & 1.00 & 0.0017 \\

Claude Opus 4.8 & Direct &
90.85 (+1.10 pp) & 80.40 & 0.241 (29.6\%) & 0.294 & 1.00 & 0.2180 \\

Grok 4.20 Reasoning & Direct &
90.68 (+0.93 pp) & 79.77 & 0.294 (36.1\%) & 0.354 & 1.00 & 0.0017 \\

Gemma 4 26B & Direct &
88.65 ($-$1.10 pp) & 75.21 & 0.461 (56.6\%) & 0.570 & 1.00 & \textbf{0.0008} \\

Gemini 3.5 Flash & Direct &
91.84 (+2.09 pp) & 82.29 & 0.734 (90.2\%) & \textbf{0.872} & 1.00 & 0.0181 \\

Gemini 3.1 Pro & Direct &
92.69 (+2.94 pp) & 84.39 & 0.735 (90.3\%) & 0.852 & 1.00 & 0.0252 \\

\midrule

Gemini 3.1 Pro & Set + Cnt + Loc &
\textbf{94.66 (+4.91 pp)} & \textbf{88.24} &
\underline{0.776 (95.3\%)} & 0.847 &
2.76 & 0.0431 \\

Gemini 3.1 Pro & Set + Cnt + Loc + Ref &
\textbf{94.66 (+4.91 pp)} & \textbf{88.24} &
\textbf{0.784 (96.3\%)} & \underline{0.857} &
3.64 & 0.0733 \\

\bottomrule
\end{tabular}}
\end{table*}

\textbf{Large-Scale Generation and SurgBox Release. }
Figure~\ref{fig:tradeoff}(b) summarizes the accuracy--cost trade-off across the evaluated configurations. The complete Gemini 3.1 Pro workflow achieves the highest localization accuracy at 0.784 mIoU and costs \$0.0733/frame on the reference set. Removing the refinement stage reduces the cost to \$0.0431/frame (41.2\% lower) while retaining 0.776 mIoU, or 99.0\% of the final workflow's localization performance. Extrapolated to all 184,498 Cholec80 frames, these rates correspond to approximately \$13.52K and \$7.95K, respectively, providing a substantially cheaper operating point when annotation budget is prioritized.

For SurgBox, we use the complete refined workflow to maximize annotation quality, producing 285,977 class-specific bounding boxes across all 80 procedures and seven instrument classes. Full generation required 683,253 MLLM calls and cost \$14,907 (\$0.0808/frame), including regeneration overhead. In comparison, the measured cold manual annotation rate implies approximately 745--1,211 hours for a single manual pass over the same corpus. At the \$7.95K cost of the non-refined workflow, manual annotation would be equally costly at only \$6.6--10.7 per hour; above this effective hourly rate, the MLLM workflow becomes less expensive. SurgBox and its associated annotation metadata are made publicly available to facilitate reuse of the generated spatial supervision.

\textbf{Downstream Detection Benchmark.}
To assess whether SurgBox provides useful supervision for conventional detectors, we construct a video-disjoint split of 56/8/16 videos for training, validation, and testing, corresponding to 128,172/16,971/39,354 frames. We benchmark eight models trained exclusively on SurgBox annotations and no human bounding boxes used for training.

Table~\ref{tab:detbench} shows that detectors trained solely from MLLM-generated supervision achieve strong performance on both the 39,354-frame held-out test set and the 197-frame expert-verified subset. YOLOv8-x reaches 0.713 mAP on the full test set, while RT-DETR-L reaches 0.839 mAP on the human-verified subset. Across models, performance against the human reference is consistently strong despite training exclusively on pseudo-labels, indicating that SurgBox provides learnable spatial supervision suitable for training standard object detectors.

\textbf{Cross-Dataset Generalization.}
We further evaluate YOLO11-m on the independently annotated CholecSeg8k benchmark using Grasper and Hook, reserving videos 28, 43, 48, 52, and 55 exclusively for testing. As shown in Table~\ref{tab:crossdata}, training only on SurgBox reaches 0.833 mAP despite being learned from a different annotation distribution, substantially outperforming the 0.623 obtained from CholecSeg8k training data drawn from the same distribution as the test set. SurgBox pretraining followed by in-domain fine-tuning further increases performance to 0.866 mAP. While the training sets differ in scale, this result highlights the central advantage of scalable MLLM supervision: large pseudo-labeled corpora can provide transferable spatial representations that remain effective under dataset shift and can be further improved with limited in-domain human annotation.

\begin{table*}[t]
\centering
\caption{Detection benchmark on the SurgBox corpus. Every detector is trained \emph{only} on MLLM pseudo-labels, with no human box in training.}
\label{tab:detbench}
\small
\setlength{\tabcolsep}{6pt}
\resizebox{\textwidth}{!}{%
\begin{tabular}{l|cccc|cccc}
\toprule
\multirow{2}{*}{Model} & \multicolumn{4}{c|}{\textbf{Test set} --- 16 held-out videos, 39{,}354 frames} & \multicolumn{4}{c}{Expert-verified subset --- 197 frames, 238 boxes} \\
\cmidrule(lr){2-5}\cmidrule(lr){6-9}
 & mAP & AP$_{50}$ & AP$_{75}$ & F1 & mAP & AP$_{50}$ & AP$_{75}$ & F1 \\
\midrule
YOLOv8-x & \textbf{0.713} & \underline{0.869} & \textbf{0.782} & \textbf{90.6} & \underline{0.833} & \underline{0.932} & \textbf{0.909} & \underline{87.8} \\
RT-DETR-L & \underline{0.710} & \underline{0.869} & \underline{0.781} & 89.8 & \textbf{0.839} & \textbf{0.937} & \underline{0.901} & \textbf{88.3} \\
YOLO11-x & 0.709 & 0.866 & 0.777 & \textbf{90.6} & 0.823 & 0.918 & 0.884 & 86.7 \\
YOLO11-m & 0.703 & 0.862 & 0.772 & \underline{90.5} & 0.811 & 0.907 & 0.887 & 86.1 \\
FCOS-R50 & 0.682 & \textbf{0.876} & 0.755 & 58.4 & 0.768 & 0.920 & 0.831 & 50.1 \\
YOLO11-n & 0.681 & 0.847 & 0.750 & 89.8 & 0.799 & 0.915 & 0.862 & 86.5 \\
RetinaNet-R50 & 0.664 & 0.866 & 0.736 & 83.8 & 0.769 & 0.916 & 0.854 & 76.4 \\
Faster R-CNN-R50 & 0.660 & 0.865 & 0.740 & 85.2 & 0.742 & 0.916 & 0.845 & 81.3 \\
\midrule
\multicolumn{9}{l}{\footnotesize
\shortstack[l]{
\textit{Training:} Ultralytics models use 640-pixel resolution, 20 epochs, COCO-pretrained initialization, and default augmentation;\\
torchvision baselines use 12 epochs under a common configuration.
}} \\
\bottomrule
\end{tabular}}
\end{table*}

\begin{table*}[t]
\centering
\caption{Cross-dataset generalisation to CholecSeg8k. Identical architecture, recipe and test set throughout. Rows differ only in the training source. }
\label{tab:crossdata}
\small
\setlength{\tabcolsep}{5pt}
\resizebox{\textwidth}{!}{%
\begin{tabular}{cl|c|ccccc|cc}
\toprule
 & \multirow{2}{*}{Training labels} & \multirow{2}{*}{Frames} & \multicolumn{5}{c|}{CholecSeg8k test (2{,}560 frames, 3{,}228 boxes)} & \multicolumn{2}{c}{AP$_{50}$} \\
\cmidrule(lr){4-8}\cmidrule(lr){9-10}
 & & & mAP & AP$_{50}$ & AP$_{75}$ & F1 & 95\% CI (F1) & Grasper & Hook \\
\midrule
A & CholecSeg8k masks \emph{(human)} & 4{,}800 & 0.623 & 0.759 & 0.714 & \underline{82.8} & [79.0, 87.5] & 0.722 & 0.524 \\
B & SurgBox pseudo-labels \emph{(none)} & 79{,}564 & \underline{0.833} & \textbf{0.973} & \underline{0.929} & \textbf{95.1} & [92.5, 97.2] & \underline{0.820} & \underline{0.847} \\
C & SurgBox $\rightarrow$ CholecSeg8k FT & 84{,}364 & \textbf{0.866} & \underline{0.972} & \textbf{0.946} & \textbf{95.1} & [93.3, 96.9] & \textbf{0.860} & \textbf{0.873} \\
\bottomrule
\end{tabular}}
\end{table*}

\subsection{Discussion}

Results suggest that scalable surgical instrument bounding-box annotation with MLLMs depends strongly on spatial grounding capability rather than semantic recognition alone. Under the same weak-label conditioning, instrument-set performance remains relatively similar across models, while localization varies widely from 0.180 to 0.735 mIoU. This separation shows that correctly identifying which instruments are present is not sufficient for reliable bounding-box generation, and that spatial grounding performance should be considered directly when selecting MLLMs for this task.

The results also show that structured decomposition can substantially improve the spatial reliability of a strong base model. For Gemini 3.1 Pro, separating set identification, instance counting, localization, and refinement increases mIoU from 0.735 to 0.784 while also improving set F1 and count accuracy, reaching 96.3\% of human--human mIoU. Most of this performance is already obtained before refinement, with the Set + Cnt + Loc configuration reaching 0.776 mIoU, suggesting that decomposition provides the main gain while refinement offers a smaller improvement in geometric precision.


\section{Conclusion}

We present a framework for converting weak frame-level instrument-presence supervision into class-specific instance-level bounding boxes and use it to construct SurgBox, covering all 184,498 annotated Cholec80 frames with 285,977 bounding boxes. The dataset, annotation workflow, and associated metadata are released to facilitate research requiring spatial supervision in laparoscopic surgery. Future work will extend the approach to additional surgical procedures and datasets to assess its generalizability beyond laparoscopic cholecystectomy. An important direction is to evaluate whether MLLMs can generate reliable bounding-box annotations without access to initial instrument-presence hints, thereby reducing dependence on existing weak supervision. Additionally, the utility of the generated bounding boxes for downstream applications could be explored, including constructing spatially grounded surgical VQA datasets and supporting instrument detection and segmentation.

\subsubsection{Acknowledgements} 
Research reported in this publication was supported by the Qatar Research Development and Innovation Council (QRDI) grant number ARG01-0522-230266. Disclaimer:  The content is solely the responsibility of the authors and does not necessarily represent the official views of Qatar Research Development and Innovation Council.

%
%
%
\bibliographystyle{splncs04}
\bibliography{ref}
%




\end{document}